# Meta-modal Information Flow: A Method for Capturing Multimodal Modular Disconnectivity in Schizophrenia

Haleh Falakshahi, Victor M. Vergara, Jingyu Liu, Daniel H. Mathalon, Judith M. Ford, James Voyvodic, Bryon A. Mueller, Aysenil Belger, Sarah McEwen, Steven G. Potkin, Adrian Preda, Hooman Rokham, Jing Sui, Jessica A. Turner, Sergey Plis, and Vince D. Calhoun

*Abstract*— *Objective:* **Multimodal measurements of the same phenomena provide complementary information and highlight different perspectives, albeit each with their own limitations. A focus on a single modality may lead to incorrect inferences, which is especially important when a studied phenomenon is a disease. In this paper, we introduce a method that takes advantage of multimodal data in addressing the hypotheses of disconnectivity and dysfunction within schizophrenia (SZ).** *Methods:* **We start with estimating and visualizing links within and among extracted multimodal data features using a Gaussian graphical model (GGM). We then propose a modularity-based method that can be applied to the GGM to identify links that are associated with mental illness across a multimodal data set. Through simulation and real data, we show our approach reveals important information about disease-related network disruptions that are missed with a focus on a single modality. We use functional MRI (fMRI), diffusion MRI (dMRI), and structural MRI (sMRI) to compute the fractional amplitude of low frequency fluctuations (fALFF), fractional anisotropy (FA), and gray matter (GM) concentration maps. These three modalities are analyzed using our modularity method.** *Results:* **Our results show missing links that are only captured by the cross-modal information that may play an important role in disconnectivity between the components.** *Conclusion:* **We identified multimodal (fALFF, FA and GM) disconnectivity in the default mode network area in patients with SZ, which would not have been detectable in a single modality.** *Significance:* **The proposed approach provides an important new tool for capturing information that is distributed among multiple imaging modalities.**

## I. INTRODUCTION

Multimodal imaging can provide useful and insightful information regarding brain health and disease [1], [2], [3], [4], [5], [6]. The motivation behind combining modalities is due to its potential of revealing hidden relationships in a set of complementary observations and discovering important variations that may unify disparate findings in brain imaging [2], [4], [7]. Multimodal techniques take advantage of complementary information from each imaging modality to provide a more comprehensive analysis of the brain and may provide a key to find missing links in complex mental illness, such as schizophrenia (SZ) [8], [9], [10]. For instance, temporal neural activity can be measured by functional magnetic resonance imaging (fMRI) [11], but it cannot provide information regarding tissue type of the brain. This is better assessed by structural MRI (sMRI) [12] and diffusion MRI (dMRI) [13]. Most previous studies analyzed each modality separately, which may disregard the multimodal cross-information [14], [15]. We have summarized previous multimodal MRI studies in SZ in the Related work section.

In this paper, we propose a method to estimate module disconnectivity from multimodal graphical models of the brain. While healthy brain graphs exhibit a modular community structure, neurodegenerative diseases such as schizophrenia may cause a breakdown of otherwise healthy communities into

This work was supported in part by NIH under grants R01EB020407, R01EB006841, P20GM103472, P30GM122734 and NSF under grant 1539067.

H. Falakshahi (hfalakshahi@gatech.edu) is with Department of Electrical and Computer Engineering, Georgia Institute of Technology (Georgia Institute of Technology, 777 Atlantic Drive NW Atlanta, GA 30332-0250) and the Tri-Institutional Center for Translational Research in Neuroimaging and Data Science (TReNDS), 55 Park Place NE, 18th Floor, Atlanta, GA 30300. V. M. Vergara is with the Tri-Institutional Center for Translational Research in Neuroimaging and Data Science (TReNDS). J. Liu is with the Tri-Institutional Center for Translational Research in Neuroimaging and Data Science (TReNDS) and Department of Computer Science, Georgia State University. D. H. Mathalon and J. M. Ford are with Department of Psychiatry, University of California. J. Voyvodic is with the Department of Radiology, Duke University. B. A. Mueller is with the Department of Psychiatry, University of Minnesota. A. Belger is with Department of Psychiatry, University of North Carolina School of Medicine. S. McEwen is with the Department of Psychiatry, University of California. S. G. Potkin and A. Preda are with the Department of Psychiatry, University of California. H. Rokham is with Department of Electrical and Computer Engineering, Georgia Institute of Technology and Tri-Institutional Center for Translational Research in Neuroimaging and Data Science (TReNDS). J. Sui is with Institute of Automation, Chinese Academy of Sciences,  University of Chinese Academy of Sciences and the Tri-Institutional Center for Translational Research in Neuroimaging and Data Science (TReNDS). J. A. Turner is with Department of Psychology, Georgia state University. S. Plis is with the Tri-Institutional Center for Translational Research in Neuroimaging and Data Science (TReNDS) and Department of Computer Science, Georgia State University. V. D. Calhoun is with the Tri-Institutional Center for Translational Research in Neuroimaging and Data Science (TReNDS) and Department of Electrical and Computer Engineering, Georgia Institute of Technology and Georgia State University, and Emory University.

small modules. Our aim is to use the proposed approach to combine and analyze three types of magnetic resonance imaging (MRI) features together to investigate connectivity alternations in SZ. We use the three-way pICA [2] approach in order to accurately estimate multimodal graph edges representing relationships among data modalities. We start with estimating and visualizing links within and among extracted multimodal data features using a Gaussian graphical model (GGM). This approach enables us to construct an interpretable graphical model that represents interaction between brain components within and between the modalities (More explanation of the three-way pICA and GGM can be found in the Theoretical background section). We then propose our modularity-based method that can be applied to estimated group graphs of patients and controls to identify links that are associated with mental illness across a multimodal data set.

The disconnection hypothesis describes SZ as a disease that disrupts the synaptic neuroplastic modulation in several brain systems [16]. This hypothesis was first laid out by Friston and Frith [17] and then followed by subsequent variants [16], [18], [19], [20]. The disconnection hypothesis has been related to both structural [21] and functional [22] brain networks. Our modularity-based method aims at finding missing links associated with disconnectivity in SZ. There are some links (edges) in the healthy control group graph that play an important role in creating contiguously connected path(s) among identifiable brain (nodes). However, the disruption of some links in the patient group graph may lead to broken paths that splits a healthy module into smaller ones in the SZ patient graph. By mimicking the structure of paths in the control group graph, we are trying to find those missing links that are associated with disconnectivity (absence of path) between some components in the patient group graph.

We apply our method to real data collected from SZ patients and a healthy group including fMRI, dMRI, and sMRI. We compute fractional amplitude of low frequency fluctuations (fALFF), fractional anisotropy (FA), and gray matter (GM) maps as input features. Using synthetic and real data, we show that our approach reveals important information about disease-related network disruptions possibly missed in analyses relying on a single modality. This approach enables us to analyze the information flow in group graphs of healthy control and SZ patient groups to identify "blocked" paths in the patient group and "missing edges" associated with the disconnectivity. A preliminary version of this work has been reported [23], [24].

The remainder of the paper is structured as follows. Section II discusses work related to multimodal MRI studies in SZ, Section III provides a theoretical background for three-way pICA and GGM; Section IV describes the details of estimating and visualizing links within and among extracted multimodal data features and introduces our modularity-based method; Section V provides the details of our results for both simulated and real data; and Section VI reviews our results and implications. We provide concluding remarks in Section VII.

## II. RELATED WORK

The number of multimodal MRI studies in SZ is still limited since each necessitates more extensive knowledge in analyzing and interpreting the outcomes in comparison with unimodal studies. In this section, we summarize the previous studies in SZ that multimodal MRI data was considered.

In [7], they analyzed data collected on a group of SZ patients and healthy controls where joint independent component analysis was used across GM and task fMRI modalities. Their findings indicated that GM group differences in bilateral parietal lobe, frontal lobe, and posterior temporal regions are associated with bilateral temporal regions activated by auditory oddball target stimuli. Subsequent researchers simultaneously explored the changes of white matter (WM) tract integrity and density in SZ using voxel-wise analyses of diffusion tensor imaging (DTI) and structural WM images [25]. Results showed abnormal WM changes mainly in the left hemisphere in patients with SZ. Further evaluation of all possible combinations of correlations from the whole brain using GM and task fMRI modalities indicated that there are stronger correlations between GM and fMRI in healthy control than patients with SZ [26]. In [27] the authors calculated separate functional and anatomical connectivity maps and then combined them for each subject. They identified group differences and a correlation with clinical symptoms by using global, regional and voxel measures and k-means network analysis. Results showed that patients with SZ had a lower anatomical connectivity and less coherence between DTI and resting fMRI. Also, within the default mode network, patients with SZ showed decoupling between structural connectivity and functional connectivity. Brain connectivity abnormalities in SZ and the relationship with behavior were further examined [28]. DTI and resting state fMRI modalities were used to assess anatomical connectivity and resting functional connectivity respectively. Using a hybrid independent component analysis (ICA) approach, anatomical and functional connectivity showed evidence of reduced connectivity in SZ patients. Another study used DTI and task fMRI [29] to investigate the relationship between altered white matter diffusivity and neural activation in patients with SZ. Results showed a significantly decreased activation in the fronto-striato-cingulate network in association with decision-making involving uncertainty in patients and increased radial diffusivity in temporal white matter. In [30] Multivariate multimodal methodology has been used to examine the linkage between cognitive biomarkers of SZ and combined information from the three MRI modalities: amplitude of low frequency fluctuations (ALFF), GM and FA using multiset canonical correlation analysis. Results showed that linked functional and structural deficits in a distributed cortico-striato-thalamic circuit can explain some aspects of cognitive impairment in patients with SZ. Interactions between fMRI contrast maps from a working memory task and dMRI data were recently studied using joint ICA [31]. This study helped elucidate our understanding of structure-function relationships in patients with SZ by characterizing linked functional and WM changes related to working memory dysfunction. A multimodal voxel-based meta-analysis was used in [32] to focus on brain regions with structural and functional abnormalities. The results of this study showed decreased GM with hyper-activation in the left inferior frontal gyrus/amygdala and decreased GM with hypo-activation in the thalamus. Thus, there is considerable evidence of multimodal brain differences in SZ patients and healthy individuals; however, there are only a few studies in the context of multimodal graphical models and there is more to be studied in this area.

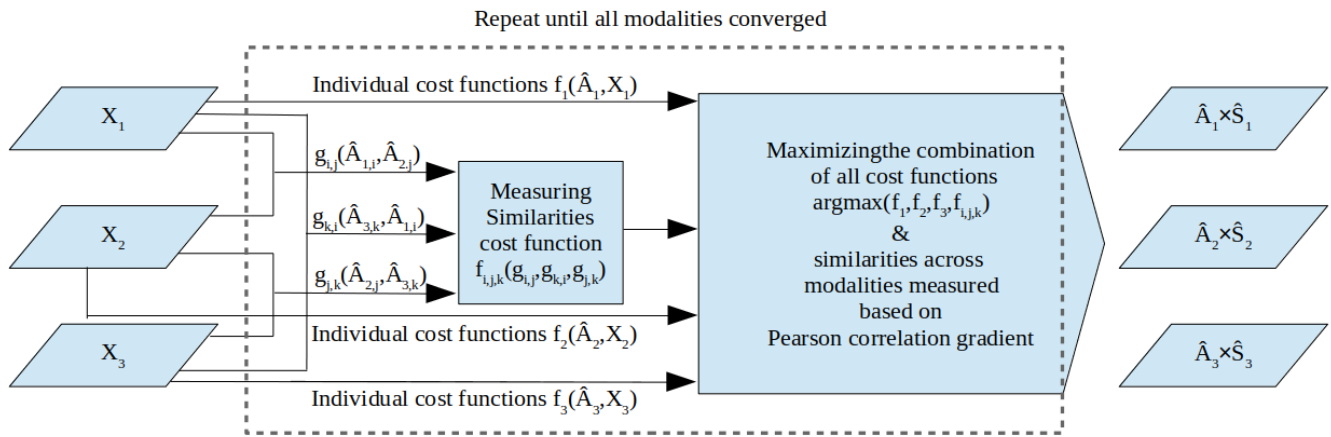

Fig. 1. Three-way pICA uses the Infomax algorithm to maximize the independence cost functions and estimates the strongest related triplet among all possible column combinations using mean statistics for triplet relationship. subsequently, similarities across modalities are measured based on Pearson correlation gradient.

## III. THEORETICAL BACKGROUND

In this section, we provide a theoretical background of the approach that we use for data fusion: three-way parallel independent component analysis (three-way pICA) [2]. Also, we explain the Gaussian graphical model (GGM) approach, which is the graphical model that we used to extract graphs of control and patient groups.

### A. Three-way pICA

The parallel ICA (pICA) is a hypothesis-free statistical technique (data-driven) that extends ICA to analyze two modalities simultaneously [33], [34], [35]. It reveals independent components from each modality and estimates the relationships between two modalities [33], [34], [35]. However, data acquisition advancement allows us to collect more than two data modalities for each subject and better take advantage of the available data [2]. In the current work, we analyze MRI data from three modalities, and we use the three-way pICA approach in order to combine them [2]. This extends the original pICA [35] approach with an ability to incorporate three modalities in one comprehensive analysis. A number of approaches for fusing data have been explored in brain imaging (see [9] for a review). The use of three-way pICA is preferred since it provides a relatively concise and straightforward way to compare the utility of the multimodal information. As Fig. 1 shows, three-way pICA algorithm seeks to maximize the statistical independence of components, while at the same time increases the correlation (of loadings from the ICAs) among modalities. Three-way pICA uses the Infomax algorithm [36], [37] to maximize the independence cost functions of three ICA factorizations: $X^{(1)} = A^{(1)}S^{(1)}$, $X^{(2)} = A^{(2)}S^{(2)}$ and $X^{(3)} = A^{(3)}S^{(3)}$ [2]. It factorizes a matrix of observations (X) for each of three features into a matrix of loading coefficients (A) and a matrix that represent statistically independent components in the measurement space (S). The algorithm seeks the strongest related triplet among all possible column combinations using mean statistics for triplet relationship and subsequently, similarities across modalities are measured based on Pearson correlation gradient [2].

### B. GGM

In this paper, we use GGM in order to construct an interpretable graphical model that represents interaction between components within and between the modalities. Since a Gaussian distribution is fully characterized by the mean vector and the covariance matrix, it suffices to determine these two quantities to construct a Gaussian model. It is more convenient to represent the covariance structure of a Gaussian model with a graph that is called a Gaussian graphical model (GGM) [38]. GGM is an undirected network of partial correlation coefficients and describes the conditional independences of multiple random variables (X1, X2, . . ., Xp) with a graph G = (V, E) where V = {1, . . ., p} as a set of nodes and E as a set of edges in which an edge between two nodes indicates they are conditionally dependent given all the other nodes [38]. The graphs that we gain from GGM are used for encoding relationships among components, wherein nodes represent components and edges that demonstrate a relationship between the connected components.

## IV. MATERIALS AND METHODS

In this section, we describe our method for meta-modal information flow analysis to determine the pathways and path-blocking features of the multimodal data. The methods consist of two parts: identifying a graph structure from multimodal data and identifying links in that structure that are likely contributors to the mental illness. We use GGM to address the former and introduce a modularity-based method for the latter. We investigate the properties of our method on synthetic data and evaluate its potential to improve our understanding of the brain and its disorders on a multimodal brain imaging data set.

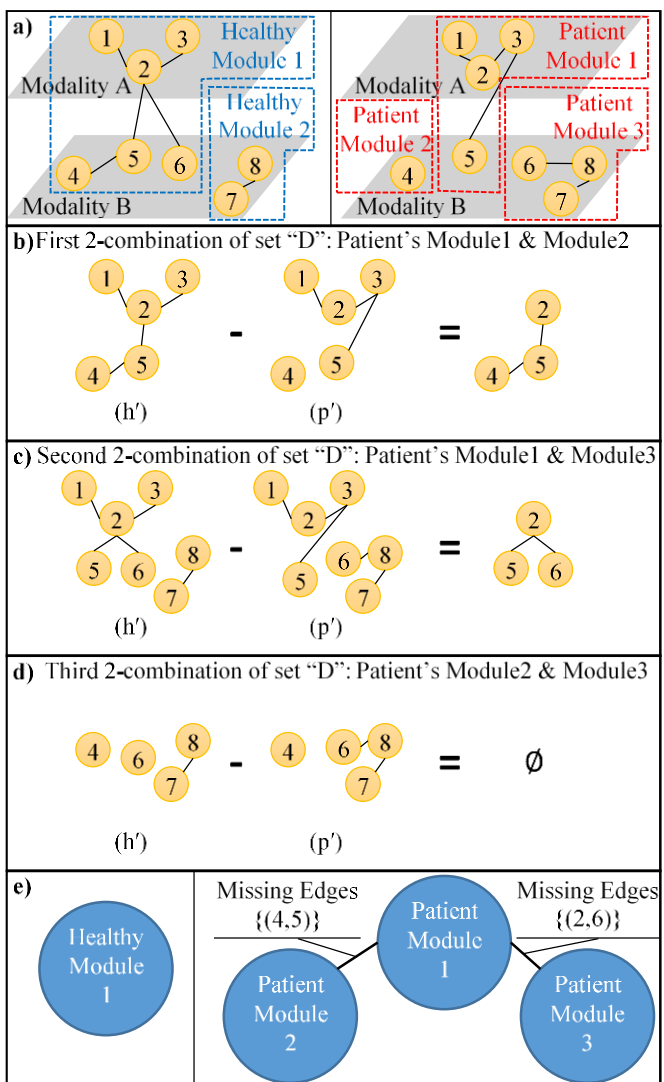

Fig. 2. Modularity Method. **(a)** Shows step 1 of the modularity method. In the first step we find the modules in the graphs of healthy (left) and patient (right) groups. When two nodes belong to the same module, it indicates that there is at least one path between them. For example, there is a path between nodes 5 and 6 in the healthy graph as they both belong to healthy_module_1. However, there is no path between nodes 5 and 6 in the patient graph since they belong to different modules. Parts **(b)**, **(c)** and **(d)** show details of step 3 of the modularity method. See text for details. Part **(e)** summarizes the results of applying the modularity method on graphs of part a. It shows that the nodes of healthy module 1 spreads into three modules in patient module 1, patient module 2, and patient module 3 and the missing edge associated with disconnectivity between nodes of module 1 and module 2 in the patient is (4,5). The missing edge associated with disconnectivity between nodes of module 1 and module 3 is indicated (2,6). Nodes of module 2 and module 3 in the patient modules indirectly have become disconnected through module 1. Missing edges associated with disconnectivity are revealed (4,5) and (2,6).

*A. Method Description*

First, we use a GGM to construct graphs of healthy and patient groups. A Gaussian distribution is completely characterized by the mean vector and the covariance matrix. Therefore, in order to construct a Gaussian model, it is adequate to specify these two quantities [39], [40].

To construct GGMs from multimodal information for healthy and patient groups, we used a joint estimation of multiple sparse Gaussian graphical model approach previously presented in [41]. The joint estimation approach merges information across classes to boost estimation of their common characteristics while retaining support for class-specific structures [42], [43]. This estimates precision matrices of the multivariate Gaussian distribution (the inverse of covariance matrices) from the observation data.

By jointly estimating precision matrices, we got adjacency matrices for the patient and healthy groups such that if we could find significant partial correlation between the two components, we will consider an edge between them. In other words, the corresponding element of them in adjacency matrix would be 1.

After constructing the graphs of healthy and patient groups, we applied our modularity method. The modularity method can identify which sets of links are likely contributing factors to mental illness. We estimated the existence or absence of paths between each pair of components in the healthy group and identified cases where a path exists in the healthy group but not in the patients by using the concept of modularity. This concept is similar to the concept of a connected graph. A graph is connected when there is a path between every pair of nodes and in a connected graph there are no unreachable nodes [44]. We first identify the modules in the healthy graph and, if those modules break into multiple modules in the patient graph, we then identify the missing edges associated with disconnectivity in the patient group.

In order to implement this idea within a modularity framework, we propose the following steps:

*1) Find all the modules in healthy graph and patient graph.*

To further clarify, consider the simple example in Fig. 2(a) that shows healthy and patient group graphs that assume just two modalities, modality A and B (left graph belongs to healthy and right graph belongs to patient). There are two modules in the healthy group graph (healthy_module_1 and healthy_module_2) and three modules in the patient group graph (patient_module_1, patient_module_2 and patient_module_3). In the first step, we find the modules in the graphs of the healthy and patient groups. Notice that when two nodes belong to the same module, it indicates that there is at least one path between them.

*2) Find disconnections that may happen in the patient group graph*

In step two, we are interested in disconnections that may occur in the patient graph. For example, in Fig. 2(a), in healthy graph (left), there is a path between node 1 and node 6 as they both belong to the same module (healthy_module_1). However, as the structure of the patient graph is different in comparison with the healthy graph (there are some additional edges and also some missing edges), node 1 and node 6 no longer belong to the same module. The reason for this is the nodes of 'healthy_module_1' spreads into multiple modules in the patient group graph. In other words, the 'healthy_module_1' breaks into multiple modules in the patient graph that reveal disconnectivity between some nodes, which means there is no path in-between anymore.

In order to detect the disconnections in the patient graph, we see if it splits into two or more modules or not for each module in the healthy group by comparing the set of nodes of the healthy module and all the patient modules nodes. If the nodes in a healthy graph module spread into two or more modules in a patient graph, this indicates a disconnectivity. Once disconnectivity is detected, the missing edges associated with this disconnectivity are obtained in step 3.

To further elucidate step 2, we go through each healthy module of the graph in Fig. 2(a). We have two healthy modules and we check both for comparison. First, we consider 'healthy_module_1'. The set of nodes of this module is {1, 2, 3, 4 ,5 ,6}. We compare this set with the sets of nodes of the patient modules:

$$V_{H1} \cap V_{P1} = \{1,2,3,5\}$$
$$V_{H1} \cap V_{P2} = \{4\}$$
$$V_{H1} \cap V_{P3} = \{6\}$$

$V_{Hi}$ denotes the nodes in the i-th healthy module and $V_{Pi}$ denotes the nodes in the i-th patient module. Because the set of nodes of 'healthy_module_1' intersects with more than one module in patient modules, this indicates disconnectivity.

We then consider 'healthy_module_2' and check if it splits into more modules in the patients or not.

$$V_{H2} \cap V_{P1} = \{\}$$
$$V_{H2} \cap V_{P2} = \{\}$$
$$V_{H2} \cap V_{P3} = \{7,8\}$$

As the set of nodes of 'healthy_module_2' has some intersection with just one module in patient, it indicates no disconnectivity.

*3) Identify missing edges associated with disconnections.*

From identified disconnectivity in step 2, we obtain the set of modules in patient group that have intersection with a module in the healthy group and put them in a set "**D**". Then from set "**D**", we consider all 2-combinations of modules and for each of them we obtain the union of their nodes and put them in set "**N**". We then induce a subgraph of the healthy graph whose nodes set is "**N**" and the edges set is the healthy graph edges where their endpoints exist in "**N**". (The two nodes forming an edge are said to be the endpoints of this edge). We named this subgraph **h′**.

We also induce a subgraph of the patient group graph whose nodes set is "**N**" and the edges set is the patient graph edges where their endpoints exist in "**N**". We named this subgraph **p′**. We then subtract the edges set of **p′** from the edges set of **h′**. (E(h′) − E(p′)). We consider those edges from the result as "disconnectors" if the endpoints belong to two different modules in the patient group graph and belong to the current healthy module we are analyzing. If E(h′) − E(p′) gives an "empty" set, this then indicates that these two modules indirectly have become disconnected through other module(s). We repeat this for all the 2-combinations of modules in set "**D**".

To describe step 3, consider the example of Fig..2; from step 2 for the graphs of Fig. 2(a), we noticed that nodes of healthy_module_1 spread into patient_module_1, patient_module_2 and patient_module_3. We put them into set "**D**". (**D** = {p_module_1, p_module_2, p_module_3}). We consider all 2-combinations modules of this set, obtain the union of their nodes and put them in set "**N**".

The first 2-combination is patient_module_1 and patient_module_2. The set "**N**" for them would be {1, 2, 3, 4, 5}. Fig. 2(b) shows subgraphs **h′** and **p′** and subtraction of the edges set of **p′** from the edges set of **h′** (E(h′) − E(p′) = {(4,5), (2,5)}). As node 4 and node 5 belong to different modules in the patient group graph, we consider this edge to be a disconnector. However, since node 2 and 5 belong to same module in the patients, we do not consider this to be a disconnector.

The second 2-combination is patient_module_1 and patient_module_3. The set "**N**" for them would be {1, 2, 3, 5, 6, 7, 8}. Fig. 2(c) shows subgraphs **h′** and **p′** in this case and subtraction of the edges set of **p′** from the edges set of **h′** (E(h′) − E(p′) = {(2,5), (2,6)}). We accept the edge (2,6) as the disconnector since its endpoints belong to two different modules in the patient graph.

The third 2-combination is patient_module_2 and patient_module_3. The set "**N**" for these would be {4, 6, 7, 8}. Fig. 2(d) shows subgraphs **h′** and **p′** and subtraction of the edges set of **p′** from the edges set of **h′**(E(h′) − E(p′) = {}). The empty set indicates that these two modules indirectly have become disconnected through other module. Fig. 2(e) summarizes the results of applying the proposed modularity method. It shows that the nodes of healthy_module_1 spreads into three modules in the patient graph (patient_module_1, patient_module_2, and patient_module_3). The missing edge associated with disconnectivity between nodes of module 1 and module 2 in patient is shown (4,5). The missing edge associated with disconnectivity between nodes of module 1 and module 3 is shown (2,6). Nodes of module 2 and module 3 in the patient graph indirectly have become disconnected through module 1. Missing edges associated with disconnectivity are identified (4,5) and (2,6).

*B. Simulation Study*

We appeal to the simulation as a proxy for method's performance as it is not possible to know the ground truth in real data.

As described earlier, we use a GGM for modeling the multi-modal information. In order to generate simulated multi-modal data, we first considered two "fixed" graph structures that assume multiple modalities including data from healthy and patient groups. The difference between the two is that the fixed graph of patient group does not include some links in comparison with the fixed graph of the healthy group. This includes missing links within or between the modalities as well as new links. We analyzed different fixed graphs (see Fig. 3 for an example). Fig. 3 (left) shows a fixed graph of a heathy group and Fig. 3 (right) shows a fixed graph for the patient group. The patient group graph is missing a few links in comparison with the fixed graph from the healthy group. For example, edge (2,5) which is missing in the patient graph and is a cross edge between modality A and modality B or edges (6,7) and (10,11) which are inside modalities B and C respectively. Also, there is an additional edge in the patient group which is (8,9). For the sake of simplicity, here in Fig. 3, we analyzed fixed graphs of simulated patient and healthy control subjects with 11 nodes. In our simulation, we generated 50 random fixed graphs using stochastic block model [45]. The simulation has three blocks of nodes with the sizes 3, 3 and 11. The edge probability within a block was set to $\frac{\ln(n)}{n}$ that it is close to the probability that a randomly chosen node in the block is a part of the largest connected component [46] where $n$ is the number of nodes in the block and the edge probabilities between the blocks were set to 0.01. We randomly generated 50 graphs for the healthy group and introduced the disconnectivity similar to Fig. 3 (right) to create another 50 graphs for the patient group. We then applied our modularity approach and generated a histogram of the number of missing links associated with disconnectivity (see Fig. 4).

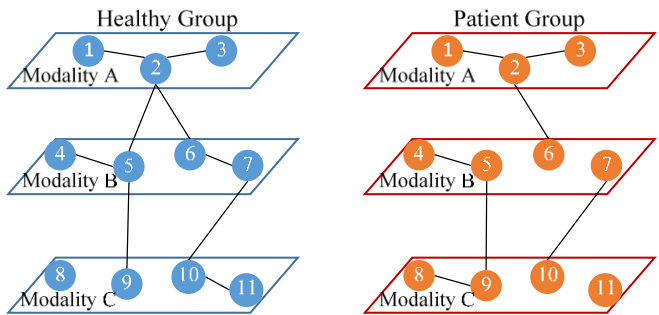

Fig. 3. Fixed graphs of a healthy group (left) and a patient group (right). The patient graph is missing some links in comparison with the fixed graph of healthy group, e.g. edge (2,5) which is an edge between modality A and modality B and edges (6,7) and (10,11) which are inside modalities B and C, respectively. Also, there is an additional edge in patient group which is (8,9).

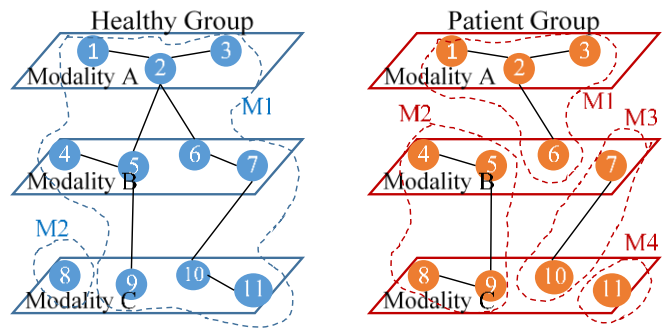

Fig. 5. Modules in the fixed graphs of a healthy group (left) and a patient group (right). There are two modules in the healthy group graph and four modules in the patient graph. Each module includes a set of nodes which there is a path between every pair of them. If two nodes belong to separate modules, it means there is no path between them.

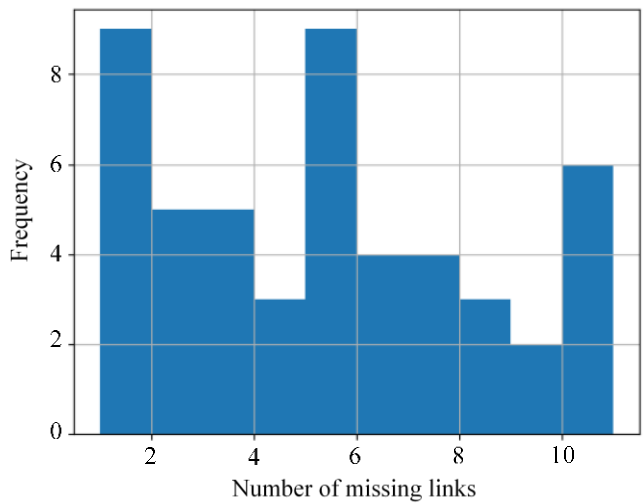

Fig. 4. Various sizes of disconnecting sets. The simulation has been applied to the different graphs' structures. The histogram of number of missing links associated with disconnectivity of random generated graphs shows that different random graphs structures were analyzed.

In order to generate a random covariance matrix, we generated a random partial correlation matrix in accordance with the structure of the aforementioned fixed graphs. The values for missing edges were uniformly randomly sampled in an interval around zero ([-0.0001 0.0001]). Values for edges were sampled from [-1, 1] interval excluding the [-0.0001 0.0001] range; while diagonal elements of the partial correlation matrix were set to 1. We then establish a precision and covariance matrices. The mean vector was set to zero. Given the covariance matrix and mean vector, we generated simulated data for each node of the healthy and patient group graphs. In order to compute the estimated graphs, we used the joint estimation of the multiple sparse Gaussian graphical model approach [41]. The simulation data was analyzed with this algorithm to obtain the estimated precision matrix for healthy and patient groups.

The mean square error of the estimated precision matrix of both groups was calculated for validation, which will be discussed in the results section. In order to identify which sets of links are likely contributing factors to mental illness, we applied the modularity approach described earlier on estimated graphs. To elucidate, consider Fig. 5 that shows modules of both groups of Fig. 3 that assumes three modalities. We have two modules in the healthy group, which means between every node of each module there is at least one path. In the patient graph, we have four modules that show disconnectivity relative to the healthy group. For example, there is a path between node 4 and node 7 in the healthy group as they both belong to a same module (module 1), but the patient graph does not have a path between them as they no longer belong to same module (e.g., node 4 belongs to module 2 and node 7 belongs to module 3 in the patient group graph). The reason for this disconnectivity is the absence of two edges: edge (2,5), which is cross edge between two modalities, and edge (6,7). By applying modularity method, we can identify all the missing links that are associated with disconnectivity. Missing links related to example of Fig. 5 will be shown in the simulation result section.

To study the effect of noise on performance of our proposed method we added different levels of white Gaussian noise to the samples from the model. By imposing different noise variances, we controlled the Signal-to-Noise Ratio (SNR). As explained earlier, we generated 50 random fixed graphs using the stochastic block model. We applied modularity method to estimate the set of disconnectors (missing links associated with disconnectivity) for each graph and kept them as the ground truth. We then have added noise to the synthetic data to evaluate model performance. We focused on model stability in terms of predicting missing edges associated with disconnectivity – the main focus of our work. For varying SNR, we estimated the appropriate level of noise variance of each node to correspond to the SNR in the -20 to 20 dB range. We used our estimator to estimate graphs of patient and control groups from the noisy data. We applied our modularity approach to these graphs and compared the outcome and the ground truth. For the sets of estimated and the ground truth disconnectors of each graph, we calculated precision (the ratio of correctly identified disconnectors to their total estimated number), recall (the ratio of correctly identified disconnectors to their total true number) and F-measure -- their harmonic average (in [0,1] range). Result are shown in the simulation results section.

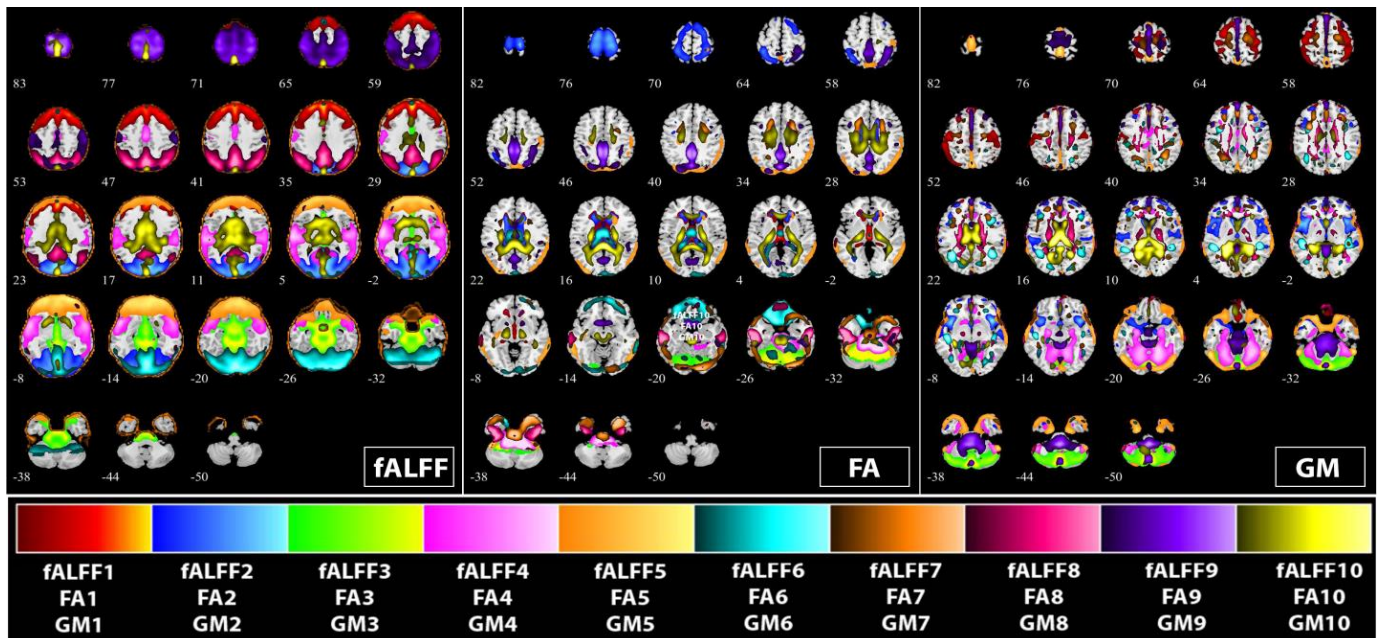

Fig. 6. Montage composite plot of components of fALFF(left), FA(middle) and GM(right). After removing the artifactual components includes one fALFF, one FA and two GM components, we have 9 components for fALFF and FA and 8 components for GM.

### C. Analysis of Multimodal Brain Imaging Data

We considered data from the fBIRN study that included fMRI, dMRI, and sMRI collected from 147 healthy subjects and 147 SZ patients [47].

The fMRI data was preprocessed using an automated analysis pipeline [48] in SMP 8. Motion correction, slice timing correction and normalization to MNI space were conducted including re-slicing to 3 × 3 × 3 mm voxels. Data spatially was smoothed with an 8 mm full width half max (FWHM) Gaussian filter. Sum of the amplitude values in the 0.01 to 0.08Hz low-frequency power range was divided by the sum of the amplitudes over the entire detectable power spectrum (range: 0–0.25Hz) to calculate fractional amplitude of low frequency fluctuations (fALFF) (see more details in [49], [50]).

The following preprocessing steps were applied to dMRI data using the FMRIB Software Library (www.fmrib.ox.ac.uk/fsl). Quality control was conducted to identify and remove excessive motion or vibration artifacts. Also, motion and eddy current correction were applied based on the correction of gradient directions for any image rotation. Diffusion tensor and scalar measures such as fractional anisotropy (FA) were calculated and smoothed using 8 mm FWHM Gaussian filter.

sMRI was normalized to MNI space using the unified segmentation method in SPM 8 and was resliced to 3 × 3 × 3 mm, and was segmented into gray matter (GM), white matter and cerebral spinal fluid (CSF). The GM images were smoothed with a FWHM of 8 mm Gaussian filter. For identifying outlier and quality control, spatial Pearson correlation with the template image was performed (details can be found in [51]).

After preprocessing, the three-dimensional brain images of each subject were reshaped into a one-dimensional vector and stacked, forming a matrix (N_subj × N_voxel) for each of the three modalities. Then these three matrices were normalized to have the same average sum of squares to ensure all modalities had the same ranges. To minimize the potential impact of gender, age and site covariates, multivariate analysis of covariance (MANCOVA) was used to adjust for these covariates prior to computing the normalized feature matrices.

After estimating fALFF, FA, and GM maps as input features, we applied three-way pICA [2] (implement in MATLAB and is part of the GIFT software) on these three features and set the number of components to 10. (see the theoretical background section for more details about three-way pICA). The output of the three-way pICA gave us 6 matrices including 3 loading matrices for fALFF, FA, and GM with the dimension of (number of subjects) × (number of component) which in our study is 294×10 and 3 components matrices with the dimension of (number of components) × (number of voxel of each feature).

The three-way pICA generated ICA components for each modality: 10 components for fALFF, 10 for FA, and 10 for GM. Relying on expert knowledge we removed artifactual components consisting of one fALFF, one FA and two GM maps. The remaining 26 components include 9 fALFF, 9 FA, and 8 GM maps. A composite montage plot of these components is shown in Fig. 6. We used the labeling tool in GIFT software (http://trendscenter.org/software/gift/) to gain the names of the regions of each component. Table 1 shows how each component is related to brain regions.

In order to extract the group graphs of healthy and patient with SZ groups, we use the joint estimation of multiple sparse Gaussian graphical model [41]. We applied a joint estimator to the 26 features of the load matrices of fALFF, FA and GM that were obtained from the three-way pICA. The outputs of the estimator are the precision matrices for healthy controls and SZ patients. From the precision matrices, we obtain the partial correlation matrices and compute the adjacency matrices of both groups by applying the test statistic for each element of partial correlation matrix for determining the significant edges. The extracted group graph is shown in the results section. After extracting the group graphs, we applied our modularity method. We next show the modules and missing edges associated with disconnectivity in the results section.

TABLE I
BRAIN REGIONS CORRESPONDING TO THE ESTIMATED COMPONENTS

| | fALLF 1 | fALLF 2 | fALLF 3 | fALLF 4 | fALLF 5 | fALLF 6 | fALLF 7 | fALLF 8 | fALLF 9 | GM 1 | GM 2 | GM 3 | GM 4 | GM 5 | GM 6 | GM 7 | GM 8 | FA 1 | FA 2 | FA 3 | FA 4 | FA 5 | FA 6 | FA 7 | FA 8 | FA 9 |
|---|---|---|---|---|---|---|---|---|---|---|---|---|---|---|---|---|---|---|---|---|---|---|---|---|---|---|
| Angular Gyrus | | | | | | | ■ | | | | | | | | | | ■ | | | | | | | | | |
| Anterior corona radiata left | | | | | | | | | | | | | | | | | | ■ | | | | | | ■ | | |
| Anterior corona radiata right, | | | | | | | | | | | | | | | | | | ■ | | | | | | | | |
| Cerebellum | | | | | | | | | | | | ■ | | | | | | | | | | | | | | |
| Cingulate Gyrus | | | | | | | ■ | | | | | | | | | | | | | | | | | | | |
| Cingulum | | | | | | | | | | | | | | | | | | | | | | | | | ■ | |
| Cuneus | | ■ | | | | | ■ | ■ | | | | | | | | | | | | | | | | | | |
| Fornix | | | | | | | | | | | | | | | | | | ■ | | | | | | | | |
| Fornix stria terminals left | | | | | | | | | | | | | | | | | | | ■ | | | | | | | |
| Fornix stria terminals right | | | | | | | | | | | | | | | | | | | ■ | | | | | | | |
| Fusiform Gyrus | | | | | | | | | | | | | ■ | | ■ | | | | | | | | | | | |
| Genu of corpus callosum | | | | | | | | | | | | | | | | | | | | | | | | ■ | | |
| Inferior Frontal Gyrus | | | | ■ | ■ | | | | | | | | ■ | | | | ■ | | | | | | | | | |
| inferior Occipital Gyrus | | | | | | | | | | | | | ■ | | | | | | | | | | | | | |
| Inferior Parietal Lobule | | | | | ■ | | | ■ | | | | | ■ | | | | ■ | | | | | | | | | |
| Insula | | | | | | ■ | | | | | | | | | | | ■ | | | | | | | | | |
| Lingual Gyrus | | ■ | | | ■ | | | | | | | | | | | | | | | | | | | | | |
| Medial Frontal Gyrus | ■ | | | ■ | | | | | | | | | | | | ■ | | | | | | | | | | |
| Middle Frontal Gyrus | ■ | | | ■ | | | | | | | | | ■ | | | | | | | | | | | | | |
| Middle Occipital Gyrus | | | ■ | | | | | | | | | | | | | | | | | | | | | | | |
| Middle Temporal Gyrus | | | ■ | | | ■ | | | | | | | | | ■ | | ■ | | | | | | | | | |
| Orbital Gyrus | | | | ■ | | | | | | | | | | | | | | | | | | | | | | |
| Para hippocampal Gyrus | | | | | ■ | ■ | | | | | | | ■ | | | ■ | ■ | | | | | | | | | |
| Paracentral Lobule | | | | | | | | | | | | | | | | ■ | | | | | | | | | | |
| Postcentral Gyrus | | | | | | | ■ | | | | | | | | | ■ | | | | | | | | | | |
| Posterior Cingulate | | ■ | | | | | ■ | | | | | | | | ■ | ■ | ■ | | | | | | | | | |
| Posterior corona radiata left | | | | | | | | | | | | | | | | | | | | | ■ | ■ | | | | |
| Posterior corona radiata right | | | | | | | | | | | | | | | | | | | | | ■ | ■ | | | | |
| Posterior limb of internal capsule left | | | | | | | | | | | | | | | | | | | | | | | | | ■ | |
| Posterior limb of internal capsule right | | | | | | | | | | | | | | | | | | | | | | | | | ■ | |
| Posterior thalamic radiation left | | | | | | | | | | | | | | | | | | | | | | | | ■ | | |
| Posterior thalamic radiation right | | | | | | | | | | | | | | | | | | | | | | | | ■ | | |
| Precentral Gyrus | | | | | | | | | | | | | ■ | | | | ■ | | | | | | | | | |
| Precuneus | | ■ | | | | | ■ | | | | | | | | ■ | ■ | ■ | | | | | | | | | |
| Rectal Gyrus | | | | ■ | | | | | | | | | | | | | | | | | | | | | | |
| Splenium of corpus callosum | | | | | | | | | | | | | | | | | | | | | ■ | ■ | | | | |
| Sub-Gyral | | | | | | | | | | | | | ■ | | | | | | | | | | | | | |
| Superior corona radiate left | | | | | | | | | | | | | | | | | | | | | | | | | | ■ |
| Superior corona radiate right | | | | | | | | | | | | | | | | | | | | | | | | | | ■ |
| Superior Frontal Gyrus | ■ | | ■ | ■ | | | | ■ | | | | | | | | ■ | | | | | | | | | | |
| Superior Parietal Lobule | | | | | | ■ | | | | | | | | | | ■ | | | | | | | | | | |
| Superior Temporal Gyrus | | | ■ | | | ■ | | | ■ | | | | | | ■ | ■ | ■ | | | | | | | | | |
| Transverse Temporal Gyrus | | | ■ | | | | | | | | | | | | | | | | | | | | | | | |
| Uncus | | | | | | | ■ | ■ | | | | | | | | | | | | | | | | | | |

To relate our synthetic experiments to the real data, we estimated the SNR in fBIRN data. The loading matrix computed by three-way parallel ICA consists of 30 components (26 networks + 4 artifact). Out of the 26 preserved networks we picked a component that is most significantly group discriminative according to the t-test. We defined SNR as the T-value for this component divided by T_fake which we got from applying t-test to data after randomly permuting the labels.

We repeated this process 1000 times each time computing the SNR. We then computed the median and the mean of all the SNRs. We discuss the result in the real data analysis results section.

V. RESULTS

A. Simulation Results

The simulation was implemented in Python 2.7 mainly using NetworkX, SciPy and Scikit-Learn libraries [52] included 1000 subjects for the healthy group and 1000 subjects for patients' group. We generated simulated data for every node (feature) of the fixed graphs displayed in Fig. 3 using GGM. After applying the joint estimation algorithm in [41] on simulated data, that was implemented in R [7], the mean square error of estimated precision matrices of both groups for 100 iterations were calculated. Estimated mean square errors are very small with the mean of 0.08. From precision matrices we gained partial correlations and performed statistical tests on the partial correlations. We considered edges between two nodes if the corresponding corrected P-value for multiple comparisons was less than a significance level of 0.05. We gained the 100 estimated graphs with 100% precision for comparison. The result of applying the proposed modularity approach on the estimated graphs is seen in Fig. 7. Missing edges associated with disconnectivity are shown in this figure as well. Module 1 in the healthy group that includes nodes 1, 2, 3, 4, 5, 6, 7, 9, 10 and 11 breaks into four modules in the patient group. For example, there is no path between the nodes of module 1 and module 2 in the patient group because of the missing edge (2,5), which is the cross edge between two modalities. Also, there is no path between nodes of module 2 and module 4 in the patient group because of missing edges {(2,5), (6,7), (10,11)}.

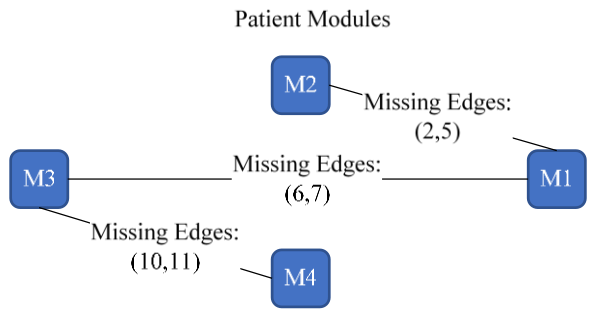

Fig. 7. Missing edges associated with disconnectivity in patient graph of Fig. 3. For instance, nodes of module2 (M2) in patient group graph that includes nodes 4, 5, 8 and 9 (according to Fig. 3) become disconnected from nodes of module1 (M1) that includes nodes 1, 2, 3 and 6. The missing edges in the patient group graph associated with this disconnectivity is (2,5).

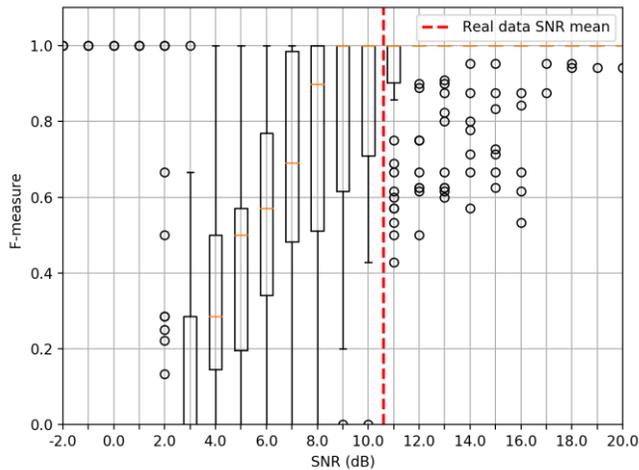

Fig. 8. Dependence of F-measure for varying SNR for synthetic data experiment (50 randomly generated graphs per SNR. To relate our synthetic experiments to the real data, we have estimated the SNR in fBIRN data red dotted line determines the mean of SNR related to real data.

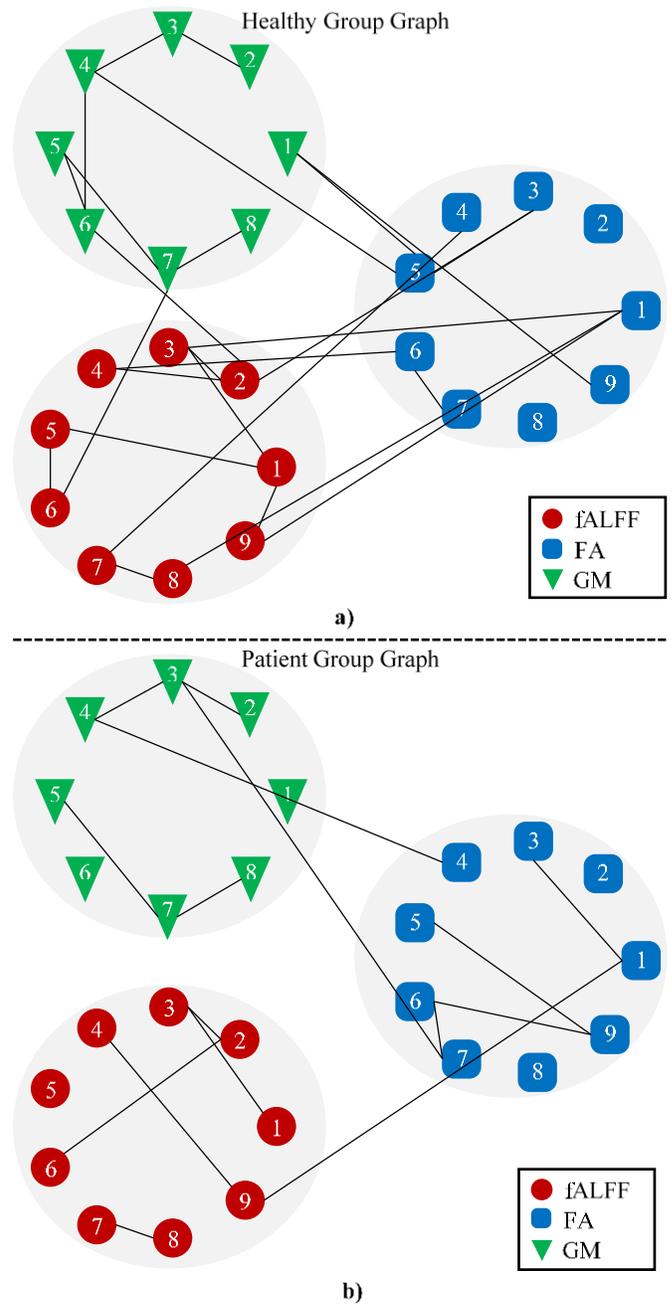

Fig. 9. Extracted group graphs of healthy group and SZ patients using GGM. After performing statistical tests on the partial correlations, if the corresponding corrected P-value for multiple comparisons was less than a significance level of 0.05, the edge was considered between two components (nodes).

As we explained in the simulation study section, we tried to investigate to what signal to noise ratio the system will work. Fig. 8 shows that our model works well when the signal to noise ratio (SNR) is higher than 7dB. It shows boxplots of f-measure and SNR of 50 synthetic random graphs.

### B. Real Data Analysis Results

After performing statistical tests on the partial correlations, we considered edges between two components if the corresponding corrected P-value for multiple comparisons was less than the significance level (0.05). The extracted group graphs of healthy and SZ patients are shown in Fig. 9.

After applying our modularity approach on extracted group graphs, we see that the healthy group graph includes three modules and the SZ patients graph includes 10 modules. This is due to one of the healthy modules splitting into eight modules in the patient group. Fig. 10(a) shows the module in the healthy group which breaks into eight modules. The Fig. 10(b) shows how the healthy module (M1) breaks into eight modules in SZ patients' graph (M1, M2, M3, M4, M5, M6, M9, M10). The modality and the ID for components of each broken module can be seen in Fig. 10(b). For example, module 4 (M4) includes components 5, 7 and 8 of GM (GM5, GM7 and GM 8). Module 2 (M2) includes components 1, 2, 3 and 6 of fALFF (fALFF1, fALFF2, fALFF3 and fALFF6). Missing edges associated with disconnectivity between the components of different modules can be seen in the Fig. 10(c). For example, the missing edge associated with disconnectivity between component GM5 located in module 4 (M4) and component fALFF1 located in module 2(M2) is (fALFF6, GM7) that is related to cross modal paths and emphasizes the importance of analyzing multimodal data in order to provide a more informed understanding of mental illness.

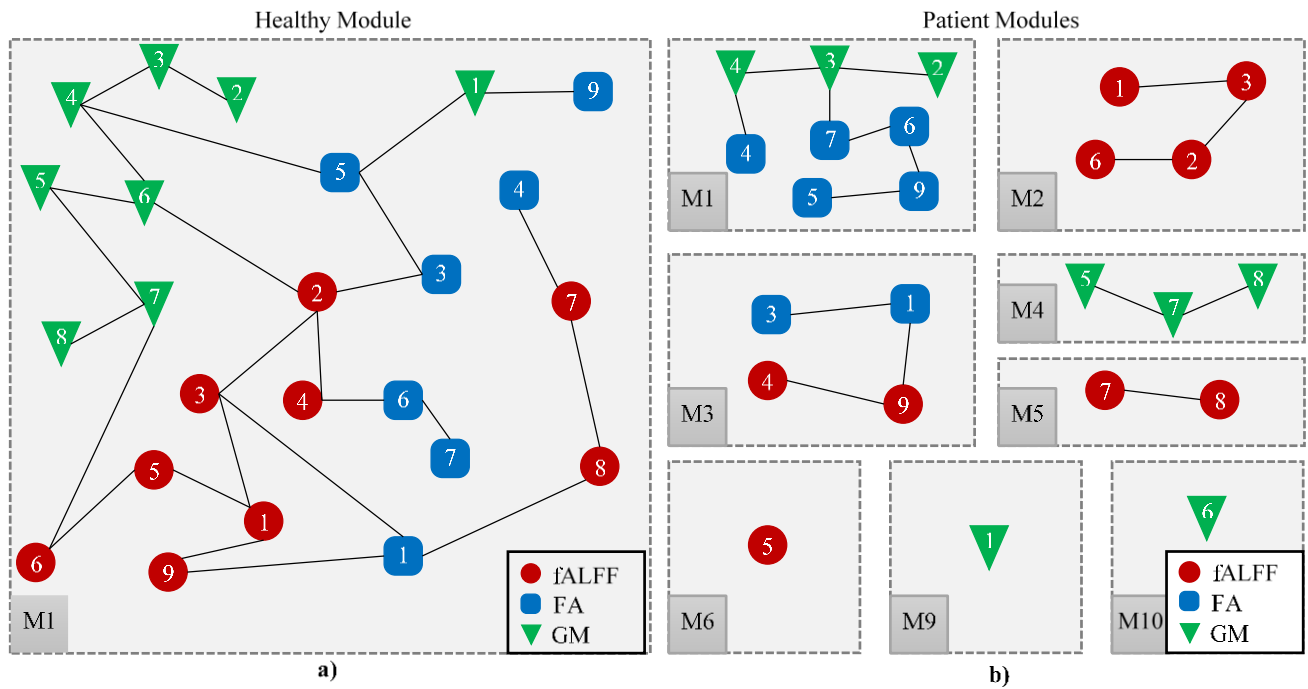

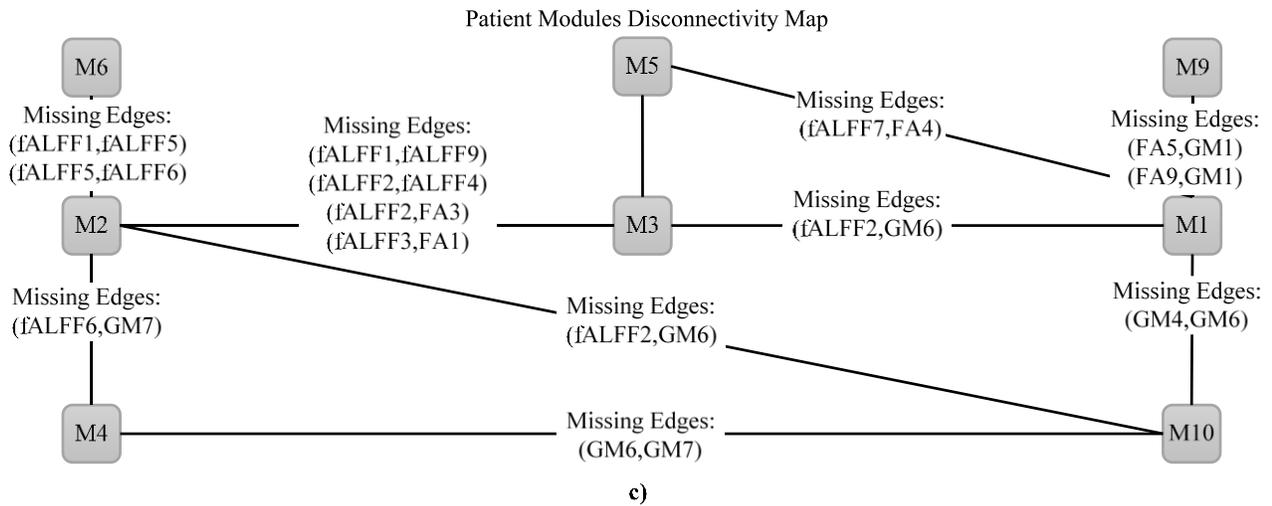

Fig. 10. (a) shows module 1 (M1) in the healthy group graph which breaks into eight (8) modules in SZ patient graph. Part (b) shows how the healthy module (M1) breaks into 8 modules in SZ patients' graph (M1, M2, M3, M4, M5, M6, M9, M10). Components of each module can be seen in part b. For example, module 4 (M4) includes components 5, 7 and 8 of GM (GM5, GM7 and GM 8). Module 2 (M2) includes components 1, 2, 3 and 6 of fALFF (fALFF1, fALFF2, fALFF3 and fALFF 6). Missing edges associated with disconnectivity between the components of different modules can be seen in part c. For example, the missing edge associated with disconnectivity between component GM5 that is located in module 4 (M4) and component ALFF 1 that is located in module 2(M2) is (fALFF6, GM7).

We investigated to see whether our model works well in real data or not by estimating the signal to noise ratio in the fBIRN data and compared it to the level of SNR that can be tolerated by our model. We computed the median and the mean of all the SNRs that we gained through the method that was explained earlier in the "analysis of multimodal brain imaging data" section. We obtained the median of 9.5 dB and the mean of 10.49 dB. One can see in Figure 8 that above these values our method is performing sufficiently well according to the F-measure. In Fig. 8 the red dotted line determines the mean of SNR related to real data.

## VI. Discussion

We proposed a modularity approach that can be applied on any undirected graph to find disconnected modules. When applied to a comparison of healthy vs. patient populations, the method finds blocked paths and missing edges in the patient graph compared to the healthy population graph. Our approach is also suitable to analyze graphs of patient and healthy groups using multimodal data. In our framework, the GGM method provides a simple but powerful approach for estimating and comparing graphs of healthy and patient groups. SZ has been shown to be associated with a disruption of the connections present in the healthy brain [7], [53], [54], [55], [56]. In applying our modularity approach to SZ and healthy control samples, we found several disconnectors that are in line with previous literature findings. Also, it can highlight the importance of considering multimodal information in gaining a better understanding of SZ through the missing edges associated with disconnections between components, which are related to cross-modal edges such as (fALFF6, GM7), (fALFF2, GM6) and (fALFF2, FA3) (see Fig. 10(c)). Our modularity method can be applied to unimodal data analysis, but as discussed in the introduction regarding the importance of multimodal analysis, we applied it to multimodal data to have a more comprehensive analysis. In this paper, we specifically focused on SZ patients, but our method can be applied to any undirected graph extracted from data related to other conditions that can be distinguished by disconnectivity since our method can easily find a disconnector.

According to our real data analysis results, one of the healthy modules that includes the majority of components spanning all three modalities (FA, GM, fALFF) breaks into eight modules in SZ. A more detailed analysis of Fig. 10, along with Fig. 6, shows that most of the identified components that fall within the default mode network (DMN) are separated in SZ patients. Module 10, module 6 and module 5 include components related to DMN and were split. The GM6 component in module 10 shows a posterior cingulate region, the fALFF5 component in module 6 includes parahippocampal gyrus, and the fALFF7 component in module 5 includes angular gyrus and cingulate gyrus regions that are typically part of the DMN that break into a disconnected module in SZ. The DMN describes a large-scale functional brain network, which is typically more active during rest periods compared to cognitively demanding tasks [57], [58] and many previous studies on DMN reported reduced default mode network connectivity in SZ patients [19], [59], [60]. New findings [61] also revealed impaired interaction among DMN subsystems in SZ patients with a reduced central role for posterior cingulate cortex (PCC) and anterior medial prefrontal cortex (aMPFC). Hence, in light of previous studies on DMN in SZ, it seems our result regarding disconnections in DMN using our modularity-based approach is consistent, but we also identify new and complex multimodal relationships with these regions.

Module 1 in the SZ group is a large module that covers much of the brain, consisting of components related to FA and GM modalities that show some integrity preserved in this module in SZ. Module 3 in SZ is a combination of several fALFF and FA components related to the sensory processing area and frontal lobe. Module 2 components include sensory processing and frontal lobe regions. These two modules show a connection between sensory and frontal regions is still preserved in the patients. These new observations can be further investigated in future works by applying our modularity based method on other data sets related to SZ.

Since SZ is a brain disorder that can be distinguished by functional disconnectivity or abnormal integration between distant brain regions [62], it is worth mentioning that additional edges in the patient group graph may create new paths that were not present in the healthy controls. This topic can be explored in future studies, whereas this paper focuses on missing edges associated with disconnections in SZ group graph. We considered unweighted undirected graphs for analysis. Additional work could focus on weighted directed networks analysis that might provide more specific information.

## VII. Conclusion

In summary, we have provided an approach to estimate and visualize links within and among multimodal data. We then proposed a modularity-based method that can identify which links are associated with mental illness across a multimodal data set that may not be achieved by separate unimodal analyses. Through simulation and application to a large SZ data set, we demonstrated that our approach reveals important information about disorder-related network disruptions that are missed in a focus on single modality. This includes components belonging to regions of the default mode network (DMN) that are separated in SZ patients. We identified missing edges between modalities and associated these with disconnectivity, which emphasizes the importance of analyzing multimodal data. Without having the multi-modal information, we are not able to identify these important missing edges in the SZ patients that play an important role in disconnectivity between the components. This highlights the utility of our approach as well as the importance of a multimodal imaging method to studying complex mental illnesses.